\documentclass[11pt]{article}

\usepackage[final]{latex/acl}

\usepackage{times}
\usepackage{latexsym}

\usepackage[T1]{fontenc}

\usepackage[utf8]{inputenc}

\usepackage{microtype}

\usepackage{inconsolata}

\usepackage{graphicx}

\usepackage[ruled,vlined, noend, linesnumbered]{algorithm2e}
\usepackage{amsmath, amsfonts}
\usepackage[most]{tcolorbox}
\usepackage{listings}
\usepackage{xcolor}
\usepackage{booktabs}
\usepackage{makecell}
\usepackage{tabularx}
\usepackage{fontawesome5}
\usetikzlibrary{positioning,arrows.meta}
\tcbuselibrary{breakable, skins, listings}

 \title{
Probing for Long-Horizon Deductive Reasoning Capabilities in \\ Language Models with Prolog
}

\author{
 \textbf{Hadeel Al-Negheimish\textsuperscript{1,2}} \hspace{1mm} \;
 \textbf{Jasna Ilieva\textsuperscript{2}} \hspace{1mm} \;
 \textbf{Yoon Kim\textsuperscript{2}}
\\
\\
 \textsuperscript{1} King Saud University,
 \textsuperscript{2} Massachusetts Institute of Technology
\\
 {
   \texttt{\{hadeelan,jasna23,yoonkim\}@mit.edu}
 }
}

\begin{document}
\maketitle
\begin{abstract}

Current frontier LLMs can theoretically process long contexts with $1M$ tokens or more. But to what extent can they go beyond simple retrieval and perform deeper reasoning over such long contexts?  We empirically investigate long-horizon reasoning capabilities of LLMs, focusing on deductive logic expressed in Prolog. We construct \textsc{ProloNg},  a synthetic testbed to probe Prolog Long Reasoning, which systematically varies the complexity (reasoning depth) of problems, where the hardest case has a reasoning depth of $22$ and $62k$ context length.  We study 8 reasoning models across 5 families of frontier LLMs, and find that performance degrades substantially as reasoning depth grows, with the majority of models approaching chance beyond depth 10.

\end{abstract}

\section{Introduction}

Logical reasoning is central to tasks where reasoning models are increasingly expected to excel, such as in mathematical reasoning and coding. Large language models (LLMs) equipped with reasoning capabilities have shown impressive performance on a range of mathematical, logical, and coding tasks \cite[inter alia]{openai2024openaio1card, Guo2025-et, yang2025qwen3technicalreport}. At what scale does the reasoning break down? We investigate this question through the lens of deductive logical reasoning, in which conclusions are derived necessarily from a set of premises by applying valid rules of inference. 

\begin{figure}[t]
    \centering
    \includegraphics[width=0.9\columnwidth]{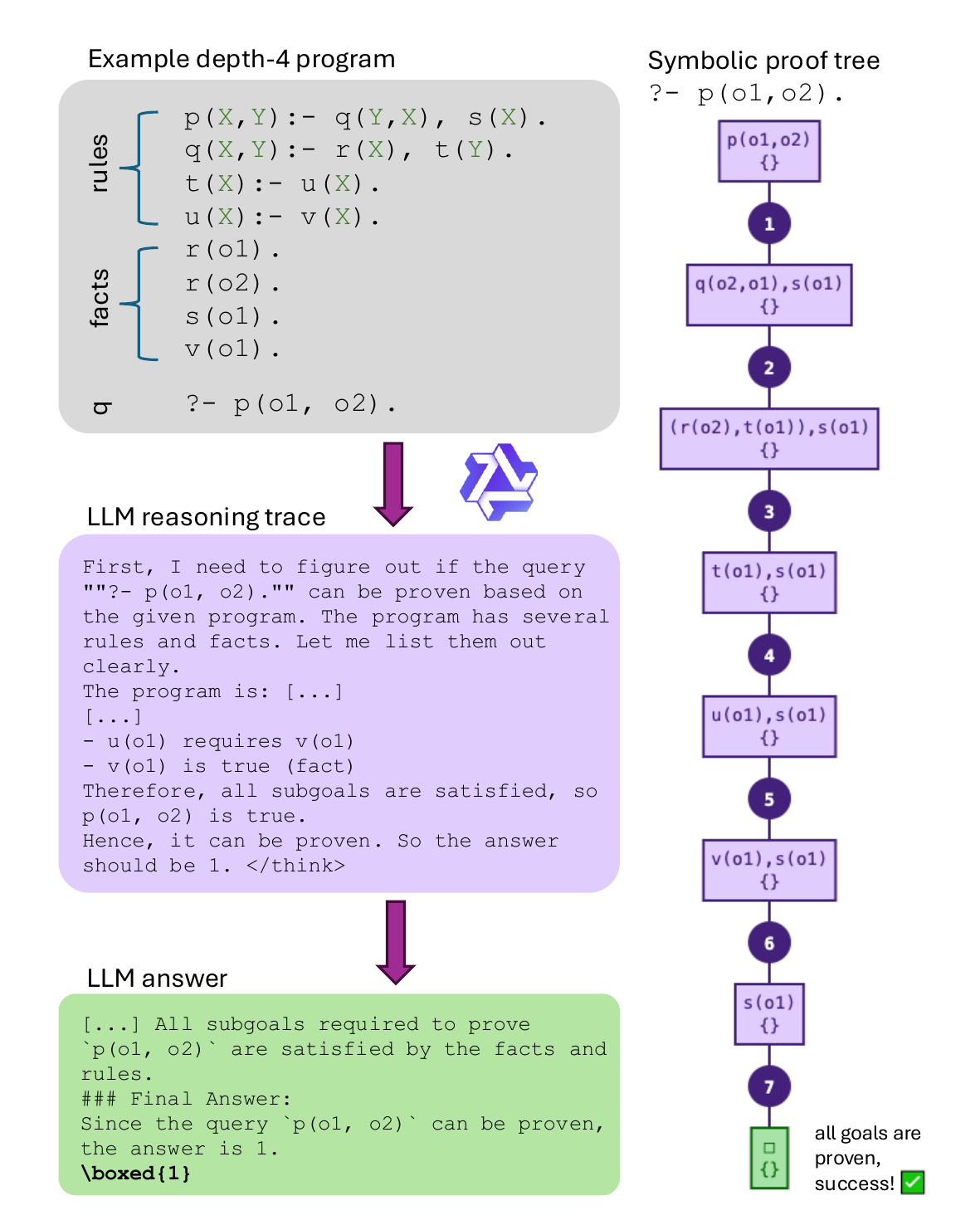}
    \caption{Example problem of depth-4, with an excerpt from the \texttt{Qwen-3-30B-Thinking} model\protect\footnotemark (left).  The symbolic derivation involves backward chaining through four rules until all subgoals are satisfied (right). See Appendix~\ref{app:policy-compliance} for an instantiation of this program as a policy compliance problem.}
    \label{fig:example}
\end{figure}

\footnotetext{Full example in Appendix \ref{appendix:ex}}

We operationalize deductive reasoning as a query-answering problem over simple Prolog-style logic programs \cite{prolog-bratko}, where the proof tree can be represented as a simple chain. As illustrated in Figure \ref{fig:example}, each problem consists of a set of rules (implications) and facts (ground literals), together with a query: can the query be proven given the program, or not? Framing this as an LLM reasoning task allows us to investigate how well LLMs can reason given explicitly specified context; do they make sound and consistent inferences? In addition to domains such as mathematics, such logical inferences are important for real-world applications, for example in policy compliance problems.\footnote{We show an instantiation of such an example in Appendix~\ref{app:policy-compliance}.}

Answering correctly requires tracing a proof tree: starting from the query and recursively resolving each subgoal against the rules until all branches terminate in facts or fail. This process is {algorithmic}; a model that has learned a robust implementation of this algorithm should, in principle, generalize beyond the shallow depths seen in previous benchmarks \cite{ijcai2020p537, tafjord-etal-2021-proofwriter, saparov2023language}.

In order to correctly answer the query, the model must correctly unify variables, follow argument orderings through rule heads and bodies (note the swapped arguments in ``\texttt{p(X,Y) :- q(Y,X), s(X).}''), and track which subgoals have and have not been established. At greater depths, this process must be sustained throughout many steps and longer contexts.

This paper systematically studies such deductive reasoning capabilities,  with depth as a primary axis of complexity, by constructing  \textsc{ProloNg}\footnote{Code and data available at \url{https://github.com/halnegheimish/ProloNg}}, a benchmark of 2,400 logic programs in Prolog with reasoning depths up to 22. Crucially, this is also a long-horizon and long-context reasoning benchmark: context length scales as the number of rules and facts in the program grows with depth, reaching a median of $62k$ tokens. 

Evaluating these problems simultaneously stresses a model's ability to reason over extended chains and to locate and use relevant information within long contexts. We perform a systematic evaluation and analysis of 8 frontier reasoning models, and find that even the best-performing frontier model, \texttt{GPT-5.5}, degrades to $67\%$ accuracy at depth-20, with the majority of models failing to perform above chance beyond depth-10. This reveals 
that even when logical inference is simple, frontier reasoning models struggle to maintain the required chain of substitutions and fact lookups as proof depth and context size scale.

\section{Related Work}

Early work on deductive reasoning benchmarks such as RuleTaker \cite{ijcai2020p537} and ProofWriter \cite{tafjord-etal-2021-proofwriter}  established the basic setup of query answering over sets of rules and facts, with ProofWriter reaching a maximum proof depth of five steps. ProntoQA \cite{saparov2023language} introduces  proof-structure controls and formal error attribution, also evaluated up to five hops. These benchmarks revealed that models at the time struggle as proof depth increases, but the depths studied remain shallow. ZebraLogic \cite{Lin-Zebra-25} identifies sharp accuracy decline in constraint-satisfaction once the search space exceeds $10^7$ possibilities.
 We show failure occurs in an even simpler setting: a smaller search space with a linear proof chain beyond depth 10. Another notable difference from previous work is that we opt to use Prolog instead of natural language in our evaluations, in order to avoid ambiguity and content effects that can impact performance, as models may draw on prior knowledge rather than the stated context \cite{10.1093/pnasnexus/pgae233}.

On the long-context processing side, LongBench-v2 \cite{bai2024longbench2}, RULER \cite{hsieh2024ruler}, and HELMET \cite{yen2025helmet} evaluate retrieval and QA over extended contexts, treating reasoning and retrieval as independent axes in most tasks. \textsc{ProloNg}  differs from these benchmarks in that every rule and fact in the program is necessary in the proof derivation, so the model cannot succeed by locating a few relevant pieces and ignoring the rest. 

\section{\textsc{ProloNg}: Prolog Long Reasoning}
\subsection{Background: Prolog}
Prolog \citep{prolog-bratko} is a declarative programming language rooted in first-order logic. A Prolog program consists of a set of facts and rules, and an interpreter evaluates the truth value of a query given the context. 
A rule, for example ``\texttt{p(X,Y) :- q(Y,X), s(X).}'' from Figure \ref{fig:example}, can be read as 
\[\forall x \forall y \left( (q(y,x)\land s(x)) \rightarrow p(x,y) \right)\]
which means that we can show $p(x,y)$ is true, if we can show that both $q(y,x)$ and $s(x)$ hold for any $x$, and $y$. The facts state unconditional truths, represented as relations between objects or object properties (predicates), ending with a period, e.g. ``\texttt{r(o1).}''. Variables are strings starting with an uppercase letter or underscore. An interpreter evaluates the truth value of a ground query, e.g. \textit{``is \texttt{p(o1,o2).} true?''} by goal-directed backward chaining through the context, and a query is determined to be true if all subgoals are satisfied, see example derivation in Figure \ref{fig:example} (right).

\subsection{Data generation}
The class of problems we consider in this framework is a smaller subset of definite Prolog logic programs, where the context can be represented as directed acyclic graphs (DAGs) (see Figure \ref{fig:context-dag} of the appendix), with no loops or recursion in rule definitions. We also limit the arity of the predicates to be of maximum two. Furthermore, each predicate is defined with only one rule; this is intended to make the reasoning trace as simple as following a single chain, with no need for backtracking or navigating unsuccessful branches. 

The main building block of the data generation process is the \textsc{GenerateRule} function (shown in Algorithm~\ref{alg:generate_rule}), which is called recursively to construct the context of each problem. We start by passing the main positive query, e.g. \texttt{p(o1,o2)}, which should be provable, along with the maximum depth and breadth, and the set of used predicate symbols. 

The \textsc{GenerateRule} function defines a rule with the target predicate as its head, and all the rules to define the body literals it includes up to the leaves. It also returns the minimum set of facts needed for the target query to succeed. An unprovable target is created by copying the set of facts, replacing them with the negative object tuple, and randomly mutating one of the facts by replacing one of the args with another constant such that the query is no longer provable. All rules and facts are then randomly shuffled. A single program (context) accompanies two samples in the data, one with a provable query and another unprovable, such that the dataset is balanced.

The algorithm by construction guarantees that at least one path in the DAG is as deep as length $N$, the other paths sampled uniformly up to that length. For each rule, we sample its breadth from a Poisson distribution with $\lambda=2.5$, and a cutoff according to $maxBreadth$. This is intended to skew the rules to having 2 body literals in order to control the size of the context. This framework allows systematic construction of random logic programs with controllable levels of complexity.

\begin{algorithm}[t]
\caption{\textsc{GenerateRule}(\textit{head}, $N$, $M$, $\mathcal{P}_u$)}
\label{alg:generate_rule}
\DontPrintSemicolon
\small
\SetKwInOut{Input}{Input}
\SetKwInOut{Output}{Output}
\SetKw{KwOr}{or}
\SetKwFunction{SampleLiteral}{SampleLiteral}
\SetKwFunction{GenerateRule}{GenerateRule}
\SetKwFunction{LiteralWithVars}{LiteralWithVars}
\Input{ground literal \textit{head}, depth $N \in \mathbb{Z}_{\ge 0}$, breadth $M \in \mathbb{Z}_{\ge 0}$, used predicate set $\mathcal{P}_u$}
\Output{Rules $\mathcal{R}$, facts $\mathcal{F}$}

\BlankLine
$\mathcal{R} \leftarrow \langle\rangle$;\quad
$\mathcal{F} \leftarrow \langle\rangle$;\quad

\BlankLine
 $U \leftarrow \textit{head.args}$\;
$\sigma \leftarrow \{U_i \mapsto x_i \mid i \in [|U|]\}$\;
$\widehat{\textit{head}} \leftarrow \LiteralWithVars(\textit{head},\, \sigma)$\;

\BlankLine
\If{$N = 0$}{
    $\mathcal{F}.\text{append}(\textit{head})$\;
    \Return $\mathcal{R},\, \mathcal{F}$\;
}

\BlankLine
\tcp{Generate body literals}
$\mathcal{B} \leftarrow \langle\rangle$\;
\For{$i \leftarrow 1$ \KwTo $M$}{
    $\ell, \mathcal{P}_u' \leftarrow \SampleLiteral(U,\, \mathcal{P}_u)$\;
    $N' \leftarrow \begin{cases} N - 1 & \text{if } i = 1 \\ \mathcal{U}(0,\, N-1) & \text{otherwise} \end{cases}$\;
    $M' \leftarrow \mathrm{Poisson}(\lambda, \textit{maxBreadth})$\;
    $(\mathcal{R}',\, \mathcal{F}') \leftarrow \GenerateRule(\ell,\, N',\, M',\, \mathcal{P}_u')$\;
    $\mathcal{R} \mathrel{+}= \mathcal{R}'$;\quad
    $\mathcal{F} \mathrel{+}= \mathcal{F}'$\;
    $\hat{\ell} \leftarrow \LiteralWithVars(\ell,\, \sigma)$\;
    $\mathcal{B}.\text{append}(\hat{\ell})$\;
}
Shuffle $\mathcal{B}$ uniformly at random\;

\BlankLine
$\mathcal{R}.\text{append}\!\left(\textit{Rule}(\widehat{\textit{head}},\, \mathcal{B})\right)$\;
\Return $\mathcal{R},\, \mathcal{F}$\;
\end{algorithm}

\subsection{\textsc{ProloNg} Benchmark}
We generate a set of 100 programs per depth, each with a provable and unprovable query, resulting in 200 samples per depth. We sample up to depth 22, skipping odd numbered $N$, in order to maximize coverage while minimizing cost. This results in a dataset of $2400$ samples at varying levels of complexity, ranging from 133 median input tokens\footnote{as measured by \texttt{DeepSeek-V4-Pro} tokenizer.} at depth 0 to $62k$ tokens at depth 22. We verify that labels are correct with the \href{https://www.swi-prolog.org/download/stable}{SWI-Prolog interpreter}, which only takes half a second even for the largest programs.
We create a canonical representation, \textit{shape},  for the program to ensure context structural diversity, which is expectedly limited in depths 0 and 2. Dataset stats are shown in Table \ref{tab:median-stats-by-depth}. We release the full dataset, generation code, and model predictions at \url{https://github.com/halnegheimish/ProloNg}.

\begin{figure}[t]
  \includegraphics[width=\columnwidth]{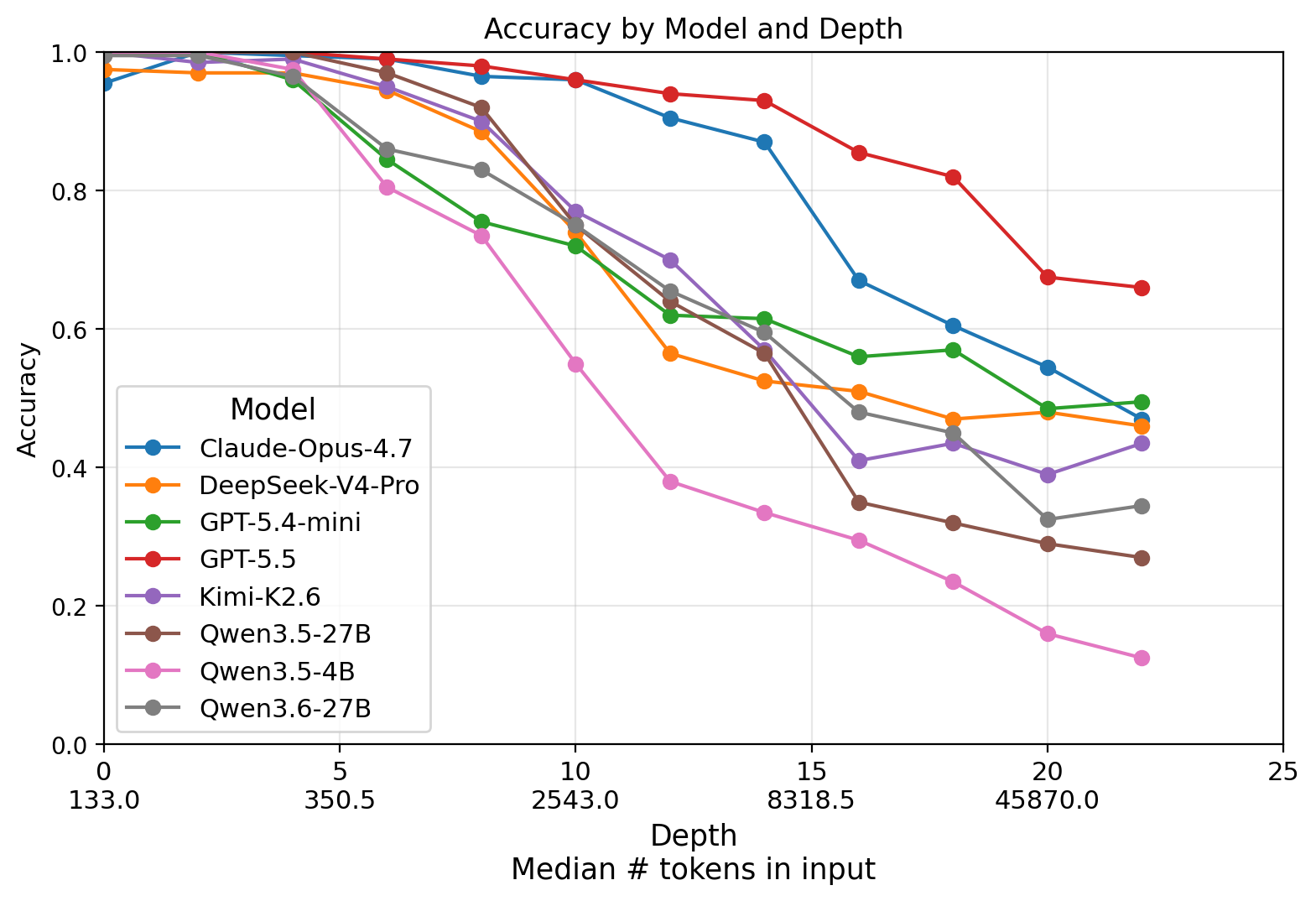}
  \caption{Model accuracy deteriorates with longer depths, with most models sharply declining after depth 10. Random chance is 0.5, since the dataset is balanced.}
  \label{fig:accuracy-per-depth}
\end{figure}
 \vspace{2mm}
\section{Results}
\subsection{Experimental setup}
\paragraph{Models} We evaluated the benchmark on frontier reasoning models, both open-weight and proprietary: \texttt{Claude-Opus-4.7}, \texttt{GPT-5.5}, \texttt{GPT-5.4-mini}, \texttt{DeepSeek-V4-Pro}, \texttt{Kimi-K2.6}, \texttt{Qwen3.6-27B}, \texttt{Qwen3.5-27B}, and \texttt{Qwen3.5-4B}. Context sizes for these models range from $256k$ tokens to $1M$, full details are shown in Table \ref{tab:model-comparison} in the appendix. 
 The models were sampled with setting the maximum output tokens to $64k$, except for \texttt{Qwen-3.5} and \texttt{3.6} models, where $82k$ tokens were recommended in the docs. When available, \texttt{reasoning effort} was set to high. 
The system prompt asks the model to output the final answer $[0|1]$ in \texttt{\textbackslash boxed\{\}}, which we extract and compare against the ground-truth.

\subsection{How does depth affect performance?}
Figure \ref{fig:accuracy-per-depth} presents accuracy as a function of reasoning depth for all evaluated models, with the median input token count annotated along the x-axis to highlight the simultaneous growth in context length. All models begin near ceiling performance at shallow depths, indicating that they are aware of the Prolog syntax. However, accuracy degrades monotonically as depth increases, with the rate and onset of decline varying considerably across models. Beyond depth 10, several models approach the chance baseline of 0.5 for this binary task. 
\texttt{GPT-5.5} is the clear outlier, maintaining high accuracy until depth 16, before declining  at depths 
$20-22$. \texttt{Claude-Opus-4.7} is the second-best performer, remaining competitive with \texttt{GPT-5.5} through depth 12 before deteriorating sharply between depths 14 and 16, ultimately settling near chance at depth 22. The remaining models follow broadly similar trajectories of steeper decline. We emphasize that the context lengths are still well within the advertised context lengths of these models.

\subsection{Models become more \textit{conservative}}
\label{sec:conservative}
\begin{figure}[t]
\centering
  \includegraphics[width=0.9\columnwidth]{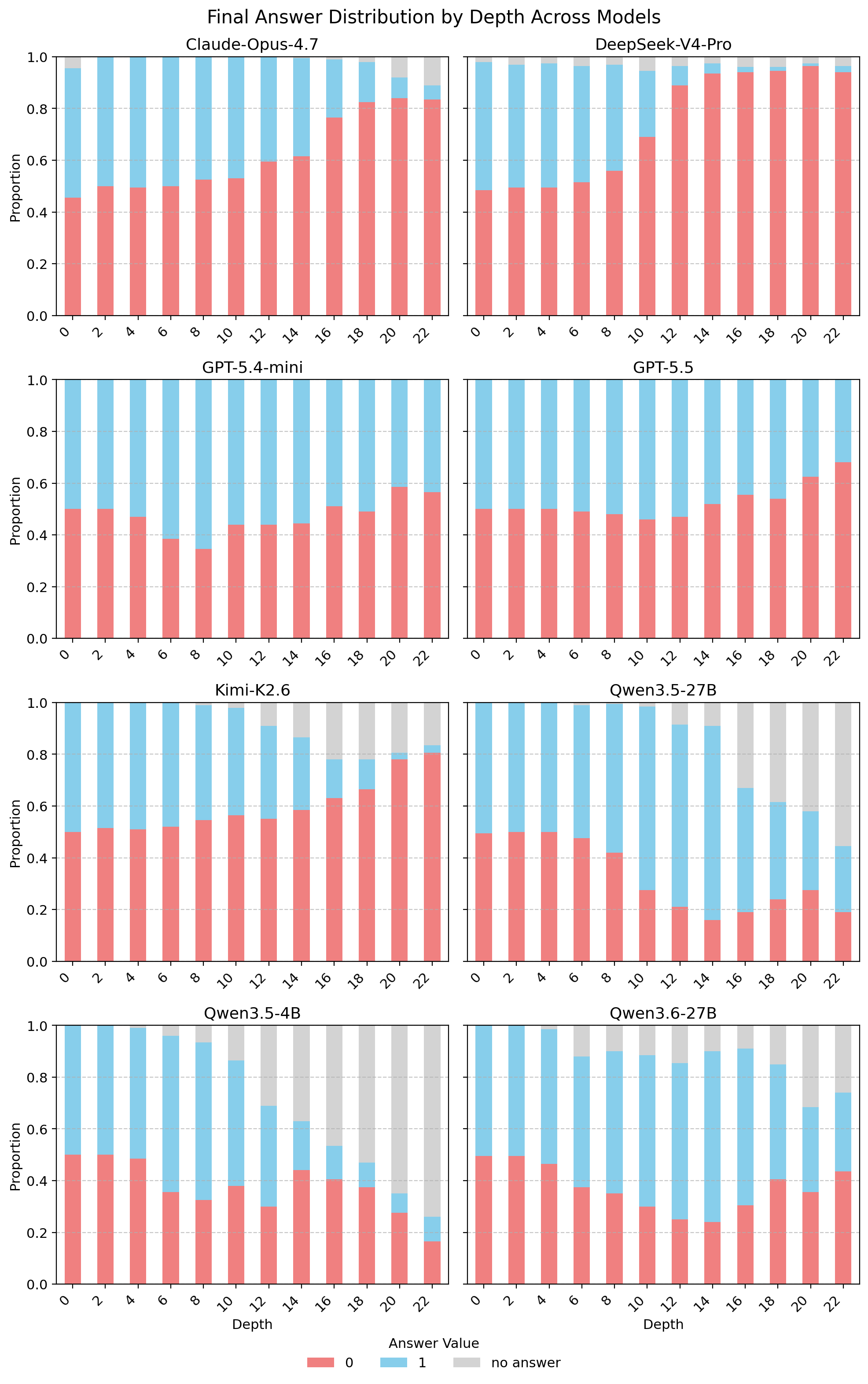}
  \caption{Proportion of extracted answers per model at each reasoning depth. Many models develop a pronounced \textit{conservative} bias, predicting that a query is unprovable (red bars), even though the dataset is balanced.}
  \label{fig:answer-freqs}
\end{figure}

We examine the frequency of final answer predictions in Figure \ref{fig:answer-freqs}. There appears to be a clear pattern in how models fail: while the benchmark is balanced by construction, with an equal number of provable/unprovable queries at every depth, many models exhibit a bias towards one answer, \textit{unprovable} (red bars), that increases with the complexity of the program. We call this the \textit{conservative} bias.

\texttt{Claude-Opus-4.7}, \texttt{DeepSeek-V4-Pro}, and \texttt{Kimi-K2.6} all develop a pronounced {conservative} bias beyond depth 10, increasingly predicting that queries are unprovable regardless of whether they actually are. By depth $16-18$, the proportion of negative predictions in these models dominates overwhelmingly. This suggests that models are effectively giving up, defaulting to ``no'' when the proof chain becomes too long to follow. \texttt{GPT-5.5} exhibits a milder version of the same tendency.
\texttt{Qwen3.5} models' main failure mode is different: they run out of tokens before reaching an answer, even at depth-10 where the median input size is only $2.5k$ tokens (grey bars). In appendix \ref{sec:trace-len}, we report the length of the reasoning traces for each model, and the \texttt{Qwen3.5} models are the only ones whose reasoning traces grow monotonically with increased depth; the rest plateau.

\subsection{What does the trace say?}
\label{sec:trace}
We investigate the reasoning trace at depth 22 of the best-performing open-source model, \texttt{DeepSeek-V4-Pro}, and the reasoning summary provided by \texttt{GPT-5.5} (best overall, but the entire trace is kept private). 
We analyze the reasoning traces of False Negatives and True Negatives. The former would tell us why it failed even when evidence to the contrary exists, and the latter would inform us whether the model correctly identified the missing fact or was right for the wrong reasons. 
Analyzing the reasoning summary produced by \texttt{GPT-5.5}, we find that in the False Negatives, the model starts the reasoning chain correctly, but then ``loses its way''. It claims that a necessary fact is missing or has no rules to support it, even though it exists in the context. We used an LLM, \texttt{GPT-5.5}, to extract the failure reason and check whether it is valid, and verified these manually for $10\%$ of the problems. For the True Negatives, LLM analysis indicates that of the 84 True Negatives, only 34 correctly cited the planted missing fact ($40.5\%$), and the vast majority did not ($59.5\%$). The other cases were: 14 only returned the final answer, without reasoning; 15 gave a vague and incomplete proof; and 21 mentioned other facts as blockers even though they are provable. This indicates that the reasoning is not faithful to the actual derivation of a query. These patterns are even more detrimental in \texttt{DeepSeek-V4-Pro} traces, where 97 of the 100 provable queries were falsely claimed to be unsupported by arbitrary conditions that exist in the context. As for the 91 true negatives, only $11\%$ of their reasoning traces correctly mention the missing fact planted.
 \vspace{2mm}
\subsection{Is it just context length?}
\label{sec:disentangle}
To better understand which factor drives models' failures, we fit a generalized linear mixed model (GLMM) \cite{bates2015lme4} with a binomial family and logit link, including model identity as a random intercept and context length and proof depth as fixed effects; full details in Appendix \ref{app:glmm}. These initial results indicate that regression restricted to the four models evaluated in both settings attributes most of the effect to context length ($\beta = -1.48$, $p < 0.001$), while proof depth appears negligible and not statistically significant ($\beta = -0.15$, $p = 0.108$). However, context length and proof depth are tightly correlated ($r= 0.97$), since, by definition, we require the context to only contain elements that would be used in the proof; the longer the proof, the longer the context. Hence, the coefficient split is highly sensitive to estimation noise rather than reflecting the two factors' true independent contributions. 

To disentangle the two factors, we regenerate the dataset such that for the same context, we evaluate the proof chain at various depths, e.g. for the context in Example~\ref{fig:example}, the original \textit{depth-4} query would be augmented with \textit{depth-}3 query \texttt{q(o2, o1)}, \textit{depth-2} \texttt{t(o1)}, and \textit{depth-1} \texttt{u(o1)}. Concretely, we generated 50 contexts at maximum depth 10, with provable and unprovable queries at each depth, at depths 0, 2, 4, 6, 8, 10, for a total of 600 samples, and evaluated the open models \texttt{DeepSeek-V4-Pro}, \texttt{Qwen3.5-4B}, \texttt{Qwen3.5-27B}, and \texttt{Qwen3.6-27B}. Since the context length is fixed across depths for each problem, there is no correlation between depth and length ($r = 0.0003$). In this setting we find that both length and depth have statistically significant and comparable impact (context-length: $\beta = -0.54$; proof-depth: $\beta = -0.36$; both $p < 0.001$). We further illustrate this effect in Table~\ref {table:accuracy-disentangle}, showing avg. accuracy for these models at each depth for both settings and their median context lengths. Models' performance on the same depth is poorer given longer context, and for the same context length, models’ performance declines with increased reasoning depth.

\begin{table}[t]
\centering
\small
\resizebox{\columnwidth}{!}{%
\begin{tabular}{rcc cc}
\toprule
& \multicolumn{2}{c}{Original} & \multicolumn{2}{c}{Shared context} \\
\cmidrule(lr){2-3}
\cmidrule(lr){4-5}
Depth & Acc & \# tokens & Acc & \# tokens \\
\midrule
0  & 0.9913 & 133.0  & 0.8550 & 2604 \\
2  & 0.9913 & 196.5  & 0.8325 & 2604 \\
4  & 0.9775 & 350.5  & 0.7725 & 2604 \\
6  & 0.8950 & 690.0  & 0.7650 & 2604 \\
8  & 0.8425 & 1328.5 & 0.7474 & 2604 \\
10 & 0.6975 & 2543.0 & 0.6825 & 2604 \\
\bottomrule
\end{tabular}
}

\caption{Accuracy and median token usage across reasoning depths for the original and shared-context settings.}
\label{table:accuracy-disentangle}
\end{table}

\vspace{2mm}

\section{Conclusion}
We present an empirical study of long-horizon deductive reasoning in frontier LLMs using Prolog. Across 8 models, accuracy collapses toward chance beyond depth 10, driven largely by a \textit{conservative bias}: models increasingly default to predicting a query is unprovable as depth grows, despite the problems remaining well within advertised context limits. Reasoning traces are often unfaithful to the actual derivation, citing missing facts or blockers that are not really there. By disentangling depth from context length in a shared-context setting, we show both factors independently and comparably degrade performance, despite being near-perfectly correlated in the original benchmark. These results suggest frontier models have not learned a robust, generalizable implementation of backward-chaining inference even in simple, linear proofs.

\clearpage
\section*{Limitations}
This framework is intended to study deductive reasoning at its simplest, so the class of generated programs does not include recursion, reasoning loops, or negation. Closed-source models do not offer access to the actual reasoning tokens used to reach the final answer, only a summary, which limits analysis of their reasoning traces.
Furthermore, running these models is computationally expensive, and API access also comes with substantial costs. Finally, the maximum number of output tokens used in this study is $64k$ for most models, and $82k$ for Qwen models, as recommended in their docs. It could be the case that increasing the budget could increase the accuracy, especially at longer depths.

\bibliography{latex/custom}

@inproceedings{Lin-Zebra-25,
author = {Lin, Bill Yuchen and Bras, Ronan Le and Richardson, Kyle and Sabharwal, Ashish and Poovendran, Radha and Clark, Peter and Choi, Yejin},
title = {ZebraLogic: on the scaling limits of LLMs for logical reasoning},
year = {2025},
booktitle = {Proceedings of the 42nd International Conference on Machine Learning},
articleno = {1507},
numpages = {17},
location = {Vancouver, Canada},
series = {ICML'25},
url={https://openreview.net/forum?id=sTAJ9QyA6l}
}

@book{prolog-bratko,
language = {eng},
author = {Bratko, Ivan},
address = {Harlow, England},
booktitle = {Prolog programming for artificial intelligence},
edition = {3rd ed.},
isbn = {0201403757},
lccn = {00026973},
publisher = {Addison Wesley},
series = {International computer science series},
title = {Prolog programming for artificial intelligence },
year = {2001},
}

@article{10.1093/pnasnexus/pgae233,
    author = {Lampinen, Andrew K and Dasgupta, Ishita and Chan, Stephanie C Y and Sheahan, Hannah R and Creswell, Antonia and Kumaran, Dharshan and McClelland, James L and Hill, Felix},
    title = {Language models, like humans, show content effects on reasoning tasks},
    journal = {PNAS Nexus},
    volume = {3},
    number = {7},
    pages = {page 233},
    year = {2024},
    month = {07},
    issn = {2752-6542},
    doi = {10.1093/pnasnexus/pgae233},
    url = {https://doi.org/10.1093/pnasnexus/pgae233},
    eprint = {https://academic.oup.com/pnasnexus/article-pdf/3/7/pgae233/58651606/pgae233.pdf},
}

@inproceedings{bai2024longbench2,
    title = "{L}ong{B}ench v2: Towards Deeper Understanding and Reasoning on Realistic Long-context Multitasks",
    author = "Bai, Yushi  and
      Tu, Shangqing  and
      Zhang, Jiajie  and
      Peng, Hao  and
      Wang, Xiaozhi  and
      Lv, Xin  and
      Cao, Shulin  and
      Xu, Jiazheng  and
      Hou, Lei  and
      Dong, Yuxiao  and
      Tang, Jie  and
      Li, Juanzi",
    editor = "Che, Wanxiang  and
      Nabende, Joyce  and
      Shutova, Ekaterina  and
      Pilehvar, Mohammad Taher",
    booktitle = "Proceedings of the 63rd Annual Meeting of the Association for Computational Linguistics (Volume 1: Long Papers)",
    month = jul,
    year = "2025",
    address = "Vienna, Austria",
    publisher = "Association for Computational Linguistics",
    url = "https://aclanthology.org/2025.acl-long.183/",
    doi = "10.18653/v1/2025.acl-long.183",
    pages = "3639--3664",
    ISBN = "979-8-89176-251-0"
}

@inproceedings{
hsieh2024ruler,
title={{RULER}: What{\textquoteright}s the Real Context Size of Your Long-Context Language Models?},
author={Cheng-Ping Hsieh and Simeng Sun and Samuel Kriman and Shantanu Acharya and Dima Rekesh and Fei Jia and Boris Ginsburg},
booktitle={First Conference on Language Modeling},
year={2024},
url={https://openreview.net/forum?id=kIoBbc76Sy}
}

@inproceedings{yen2025helmet,
  title={HELMET: How to Evaluate Long-Context Language Models Effectively and Thoroughly}, 
  author={Howard Yen and Tianyu Gao and Minmin Hou and Ke Ding and Daniel Fleischer and Peter Izsak and Moshe Wasserblat and Danqi Chen},
  year={2025},
  booktitle={International Conference on Learning Representations (ICLR)},
  url={https://openreview.net/forum?id=293V3bJbmE}
}

@inproceedings{
saparov2023language,
title={Language Models Are Greedy Reasoners: A Systematic Formal Analysis of Chain-of-Thought},
author={Abulhair Saparov and He He},
booktitle={International Conference on Learning Representations (ICLR)},
year={2023},
url={https://openreview.net/forum?id=qFVVBzXxR2V}
}

@inproceedings{tafjord-etal-2021-proofwriter,
    title = "{P}roof{W}riter: Generating Implications, Proofs, and Abductive Statements over Natural Language",
    author = "Tafjord, Oyvind  and
      Dalvi, Bhavana  and
      Clark, Peter",
    editor = "Zong, Chengqing  and
      Xia, Fei  and
      Li, Wenjie  and
      Navigli, Roberto",
    booktitle = "Findings of the Association for Computational Linguistics: ACL-IJCNLP 2021",
    month = aug,
    year = "2021",
    address = "Online",
    publisher = "Association for Computational Linguistics",
    url = "https://aclanthology.org/2021.findings-acl.317/",
    doi = "10.18653/v1/2021.findings-acl.317",
    pages = "3621--3634"
}

@inproceedings{ijcai2020p537,
  title     = {Transformers as Soft Reasoners over Language},
  author    = {Clark, Peter and Tafjord, Oyvind and Richardson, Kyle},
  booktitle = {Proceedings of the Twenty-Ninth International Joint Conference on
               Artificial Intelligence, {IJCAI-20}},
  publisher = {International Joint Conferences on Artificial Intelligence Organization},
  editor    = {Christian Bessiere},
  pages     = {3882--3890},
  year      = {2020},
  month     = {7},
  note      = {Main track},
  doi       = {10.24963/ijcai.2020/537},
  url       = {https://doi.org/10.24963/ijcai.2020/537},
}

@ARTICLE{Guo2025-et,
  title     = "{DeepSeek}-{R1} incentivizes reasoning in {LLMs} through
               reinforcement learning",
  author    = "Guo, Daya and Yang, Dejian and Zhang, Haowei and Song, Junxiao
               and Wang, Peiyi and Zhu, Qihao and Xu, Runxin and Zhang, Ruoyu
               and Ma, Shirong and Bi, Xiao and Zhang, Xiaokang and Yu, Xingkai
               and Wu, Yu and Wu, Z F and Gou, Zhibin and Shao, Zhihong and Li,
               Zhuoshu and Gao, Ziyi and Liu, Aixin and Xue, Bing and Wang,
               Bingxuan and Wu, Bochao and Feng, Bei and Lu, Chengda and Zhao,
               Chenggang and Deng, Chengqi and Ruan, Chong and Dai, Damai and
               Chen, Deli and Ji, Dongjie and Li, Erhang and Lin, Fangyun and
               Dai, Fucong and Luo, Fuli and Hao, Guangbo and Chen, Guanting and
               Li, Guowei and Zhang, H and Xu, Hanwei and Ding, Honghui and Gao,
               Huazuo and Qu, Hui and Li, Hui and Guo, Jianzhong and Li, Jiashi
               and Chen, Jingchang and Yuan, Jingyang and Tu, Jinhao and Qiu,
               Junjie and Li, Junlong and Cai, J L and Ni, Jiaqi and Liang, Jian
               and Chen, Jin and Dong, Kai and Hu, Kai and You, Kaichao and Gao,
               Kaige and Guan, Kang and Huang, Kexin and Yu, Kuai and Wang, Lean
               and Zhang, Lecong and Zhao, Liang and Wang, Litong and Zhang,
               Liyue and Xu, Lei and Xia, Leyi and Zhang, Mingchuan and Zhang,
               Minghua and Tang, Minghui and Zhou, Mingxu and Li, Meng and Wang,
               Miaojun and Li, Mingming and Tian, Ning and Huang, Panpan and
               Zhang, Peng and Wang, Qiancheng and Chen, Qinyu and Du, Qiushi
               and Ge, Ruiqi and Zhang, Ruisong and Pan, Ruizhe and Wang, Runji
               and Chen, R J and Jin, R L and Chen, Ruyi and Lu, Shanghao and
               Zhou, Shangyan and Chen, Shanhuang and Ye, Shengfeng and Wang,
               Shiyu and Yu, Shuiping and Zhou, Shunfeng and Pan, Shuting and
               Li, S S and Zhou, Shuang and Wu, Shaoqing and Yun, Tao and Pei,
               Tian and Sun, Tianyu and Wang, T and Zeng, Wangding and Liu, Wen
               and Liang, Wenfeng and Gao, Wenjun and Yu, Wenqin and Zhang,
               Wentao and Xiao, W L and An, Wei and Liu, Xiaodong and Wang,
               Xiaohan and Chen, Xiaokang and Nie, Xiaotao and Cheng, Xin and
               Liu, Xin and Xie, Xin and Liu, Xingchao and Yang, Xinyu and Li,
               Xinyuan and Su, Xuecheng and Lin, Xuheng and Li, X Q and Jin,
               Xiangyue and Shen, Xiaojin and Chen, Xiaosha and Sun, Xiaowen and
               Wang, Xiaoxiang and Song, Xinnan and Zhou, Xinyi and Wang, Xianzu
               and Shan, Xinxia and Li, Y K and Wang, Y Q and Wei, Y X and
               Zhang, Yang and Xu, Yanhong and Li, Yao and Zhao, Yao and Sun,
               Yaofeng and Wang, Yaohui and Yu, Yi and Zhang, Yichao and Shi,
               Yifan and Xiong, Yiliang and He, Ying and Piao, Yishi and Wang,
               Yisong and Tan, Yixuan and Ma, Yiyang and Liu, Yiyuan and Guo,
               Yongqiang and Ou, Yuan and Wang, Yuduan and Gong, Yue and Zou,
               Yuheng and He, Yujia and Xiong, Yunfan and Luo, Yuxiang and You,
               Yuxiang and Liu, Yuxuan and Zhou, Yuyang and Zhu, Y X and Huang,
               Yanping and Li, Yaohui and Zheng, Yi and Zhu, Yuchen and Ma,
               Yunxian and Tang, Ying and Zha, Yukun and Yan, Yuting and Ren, Z
               Z and Ren, Zehui and Sha, Zhangli and Fu, Zhe and Xu, Zhean and
               Xie, Zhenda and Zhang, Zhengyan and Hao, Zhewen and Ma, Zhicheng
               and Yan, Zhigang and Wu, Zhiyu and Gu, Zihui and Zhu, Zijia and
               Liu, Zijun and Li, Zilin and Xie, Ziwei and Song, Ziyang and Pan,
               Zizheng and Huang, Zhen and Xu, Zhipeng and Zhang, Zhongyu and
               Zhang, Zhen",
  journal   = "Nature",
  publisher = "Springer Science and Business Media LLC",
  volume    =  645,
  number    =  8081,
  pages     = "633--638",
  url = "https://doi.org/10.1038/s41586-025-09422-z",
  month     =  sep,
  year      =  2025,
  language  = "en"
}

@misc{openai2024openaio1card,
      title={OpenAI o1 System Card}, 
      author={Aaron Jaech and Adam Kalai and Adam Lerer and Adam Richardson and Ahmed El-Kishky and Aiden Low and Alec Helyar and Aleksander Madry and Alex Beutel and Alex Carney and Alex Iftimie and Alex Karpenko and Alex Tachard Passos and Alexander Neitz and Alexander Prokofiev and Alexander Wei and Allison Tam and Ally Bennett and Ananya Kumar and Andre Saraiva and Andrea Vallone and Andrew Duberstein and Andrew Kondrich and Andrey Mishchenko and Andy Applebaum and Angela Jiang and Ashvin Nair and Barret Zoph and Behrooz Ghorbani and Bohan Zhang and Ben Rossen and Benjamin Sokolowsky and Boaz Barak and Bob McGrew and Borys Minaiev and Botao Hao and Bowen Baker and Brandon Houghton and Brandon McKinzie and Brydon Eastman and Camillo Lugaresi and Cary Bassin and Cary Hudson and Chak Ming Li and Charles de Bourcy and Chelsea Voss and Chen Shen and Chong Zhang and Chris Koch and Chris Orsinger and Christopher Hesse and Claudia Fischer and Clive Chan and Dan Roberts and Daniel Kappler and Daniel Levy and Daniel Selsam and David Dohan and David Farhi and David Mely and David Robinson and Dimitris Tsipras and Doug Li and Dragos Oprica and Eben Freeman and Eddie Zhang and Edmund Wong and Elizabeth Proehl and Enoch Cheung and Eric Mitchell and Eric Wallace and Erik Ritter and Evan Mays and Fan Wang and Felipe Petroski Such and Filippo Raso and Florencia Leoni and Foivos Tsimpourlas and Francis Song and Fred von Lohmann and Freddie Sulit and Geoff Salmon and Giambattista Parascandolo and Gildas Chabot and Grace Zhao and Greg Brockman and Guillaume Leclerc and Hadi Salman and Haiming Bao and Hao Sheng and Hart Andrin and Hessam Bagherinezhad and Hongyu Ren and Hunter Lightman and Hyung Won Chung and Ian Kivlichan and Ian O'Connell and Ian Osband and Ignasi Clavera Gilaberte and Ilge Akkaya and Ilya Kostrikov and Ilya Sutskever and Irina Kofman and Jakub Pachocki and James Lennon and Jason Wei and Jean Harb and Jerry Twore and Jiacheng Feng and Jiahui Yu and Jiayi Weng and Jie Tang and Jieqi Yu and Joaquin Quiñonero Candela and Joe Palermo and Joel Parish and Johannes Heidecke and John Hallman and John Rizzo and Jonathan Gordon and Jonathan Uesato and Jonathan Ward and Joost Huizinga and Julie Wang and Kai Chen and Kai Xiao and Karan Singhal and Karina Nguyen and Karl Cobbe and Katy Shi and Kayla Wood and Kendra Rimbach and Keren Gu-Lemberg and Kevin Liu and Kevin Lu and Kevin Stone and Kevin Yu and Lama Ahmad and Lauren Yang and Leo Liu and Leon Maksin and Leyton Ho and Liam Fedus and Lilian Weng and Linden Li and Lindsay McCallum and Lindsey Held and Lorenz Kuhn and Lukas Kondraciuk and Lukasz Kaiser and Luke Metz and Madelaine Boyd and Maja Trebacz and Manas Joglekar and Mark Chen and Marko Tintor and Mason Meyer and Matt Jones and Matt Kaufer and Max Schwarzer and Meghan Shah and Mehmet Yatbaz and Melody Y. Guan and Mengyuan Xu and Mengyuan Yan and Mia Glaese and Mianna Chen and Michael Lampe and Michael Malek and Michele Wang and Michelle Fradin and Mike McClay and Mikhail Pavlov and Miles Wang and Mingxuan Wang and Mira Murati and Mo Bavarian and Mostafa Rohaninejad and Nat McAleese and Neil Chowdhury and Neil Chowdhury and Nick Ryder and Nikolas Tezak and Noam Brown and Ofir Nachum and Oleg Boiko and Oleg Murk and Olivia Watkins and Patrick Chao and Paul Ashbourne and Pavel Izmailov and Peter Zhokhov and Rachel Dias and Rahul Arora and Randall Lin and Rapha Gontijo Lopes and Raz Gaon and Reah Miyara and Reimar Leike and Renny Hwang and Rhythm Garg and Robin Brown and Roshan James and Rui Shu and Ryan Cheu and Ryan Greene and Saachi Jain and Sam Altman and Sam Toizer and Sam Toyer and Samuel Miserendino and Sandhini Agarwal and Santiago Hernandez and Sasha Baker and Scott McKinney and Scottie Yan and Shengjia Zhao and Shengli Hu and Shibani Santurkar and Shraman Ray Chaudhuri and Shuyuan Zhang and Siyuan Fu and Spencer Papay and Steph Lin and Suchir Balaji and Suvansh Sanjeev and Szymon Sidor and Tal Broda and Aidan Clark and Tao Wang and Taylor Gordon and Ted Sanders and Tejal Patwardhan and Thibault Sottiaux and Thomas Degry and Thomas Dimson and Tianhao Zheng and Timur Garipov and Tom Stasi and Trapit Bansal and Trevor Creech and Troy Peterson and Tyna Eloundou and Valerie Qi and Vineet Kosaraju and Vinnie Monaco and Vitchyr Pong and Vlad Fomenko and Weiyi Zheng and Wenda Zhou and Wenting Zhan and Wes McCabe and Wojciech Zaremba and Yann Dubois and Yinghai Lu and Yining Chen and Young Cha and Yu Bai and Yuchen He and Yuchen Zhang and Yunyun Wang and Zheng Shao and Zhuohan Li},
      year={2024},
      eprint={2412.16720},
      archivePrefix={arXiv},
      primaryClass={cs.AI},
      url={https://arxiv.org/abs/2412.16720}, 
}

@misc{yang2025qwen3technicalreport,
      title={Qwen3 Technical Report}, 
      author={An Yang and Anfeng Li and Baosong Yang and Beichen Zhang and Binyuan Hui and Bo Zheng and Bowen Yu and Chang Gao and Chengen Huang and Chenxu Lv and Chujie Zheng and Dayiheng Liu and Fan Zhou and Fei Huang and Feng Hu and Hao Ge and Haoran Wei and Huan Lin and Jialong Tang and Jian Yang and Jianhong Tu and Jianwei Zhang and Jianxin Yang and Jiaxi Yang and Jing Zhou and Jingren Zhou and Junyang Lin and Kai Dang and Keqin Bao and Kexin Yang and Le Yu and Lianghao Deng and Mei Li and Mingfeng Xue and Mingze Li and Pei Zhang and Peng Wang and Qin Zhu and Rui Men and Ruize Gao and Shixuan Liu and Shuang Luo and Tianhao Li and Tianyi Tang and Wenbiao Yin and Xingzhang Ren and Xinyu Wang and Xinyu Zhang and Xuancheng Ren and Yang Fan and Yang Su and Yichang Zhang and Yinger Zhang and Yu Wan and Yuqiong Liu and Zekun Wang and Zeyu Cui and Zhenru Zhang and Zhipeng Zhou and Zihan Qiu},
      year={2025},
      eprint={2505.09388},
      archivePrefix={arXiv},
      primaryClass={cs.CL},
      url={https://arxiv.org/abs/2505.09388}, 
}

@Article{bates2015lme4,
    title = {Fitting Linear Mixed-Effects Models Using {lme4}},
    author = {Douglas Bates and Martin M{\"a}chler and Ben Bolker and
      Steve Walker},
    journal = {Journal of Statistical Software},
    year = {2015},
    volume = {67},
    number = {1},
    pages = {1--48},
    doi = {10.18637/jss.v067.i01},
  }

\appendix
\section{Models}
 We evaluated frontier reasoning models, both open-weight and proprietary: \texttt{Claude-Opus-4.7}, \texttt{GPT-5.5}, \texttt{GPT-5.4-mini}, \texttt{DeepSeek-V4-Pro}, \texttt{Kimi-K2.6}, \texttt{Qwen3.6-27B}, \texttt{Qwen3.5-27B}, and \texttt{Qwen3.5-4B}. Full details are in Table~
 \ref{tab:model-comparison}.
\begin{table*}[ht]
\centering
\small
\begin{tabularx}{\textwidth}{p{2.8cm} p{1.3cm} p{2.5cm} p{2.4cm} X}
\toprule
\textbf{Model name} &
\textbf{Weights} &
\textbf{Model size} &
\textbf{Release date} &
\textbf{Context size} \\
\midrule

\texttt{Claude-Opus-4.7} &
\faLock &
Not disclosed &
Apr. 16, 2026 &
$1M$ tokens \\

\texttt{GPT-5.5} &
\faLock &
Not disclosed &
Apr. 23, 2026 &
$1M$ tokens \\
\texttt{GPT-5.4-mini} &
\faLock &
Not disclosed &
Mar. 17, 2026 &
$400k$ tokens \\

\texttt{DeepSeek-V4-Pro} &
\faKey &
$1.6T$ ($49B$ activated) &
Apr. 24, 2026 &
$1M$ tokens \\

\texttt{Kimi-K2.6} &
\faKey &
 $1T$ ($32B$ activated) &
Apr. 20, 2026 &
$256k$ tokens \\

\texttt{Qwen3.6-27B} &
\faKey &
$27B$ &
Apr. 21, 2026 &
$262k$ native; up to $\sim1M$ extended \\
\texttt{Qwen3.5-27B} &
\faKey &
$27B$ &
Feb. 24, 2026 &
$262k$ native; up to $\sim1M$ extended \\
\texttt{Qwen3.5-4B} &
\faKey &
$4B$ &
Feb. 27, 2026 &
$262k$ native; up to $\sim1M$ extended \\

\bottomrule
\end{tabularx}
\caption{Comparison of frontier models used in this analysis based on publicly available data.}
\label{tab:model-comparison}
\end{table*}

\section{Practical Example}
\label{app:policy-compliance}
The same program as in Figure~\ref{fig:example} can be instantiated as a practical example of policy compliance, as shown below: 
\begin{tcolorbox}[
  colback=gray!5!white,
  colframe=gray!55!black,
  title=\textbf{Policy Compliance Example},
  fonttitle=\bfseries,
  breakable
]
\begin{lstlisting}
% An employee may access a system if:
% 1. they are approved for the system, and
% 2. they completed security training.

may_access(Employee, System) :-
    approved(System, Employee),
    trained(Employee).

% Approval requires the system to be authorized
% and the employee to be verified.
approved(System, Employee) :-
    authorized_system(System),
    verified(Employee).

verified(Employee) :-
    background_checked(Employee).

background_checked(Employee) :-
    identity_confirmed(Employee).

authorized_system(payroll).
authorized_system(hr_portal).

trained(alice).
identity_confirmed(alice).

?- may_access(alice, hr_portal).
\end{lstlisting}
\end{tcolorbox}
\section{\textsc{ProloNg} Data}
\begin{figure}[t]
  \includegraphics[width=\columnwidth]{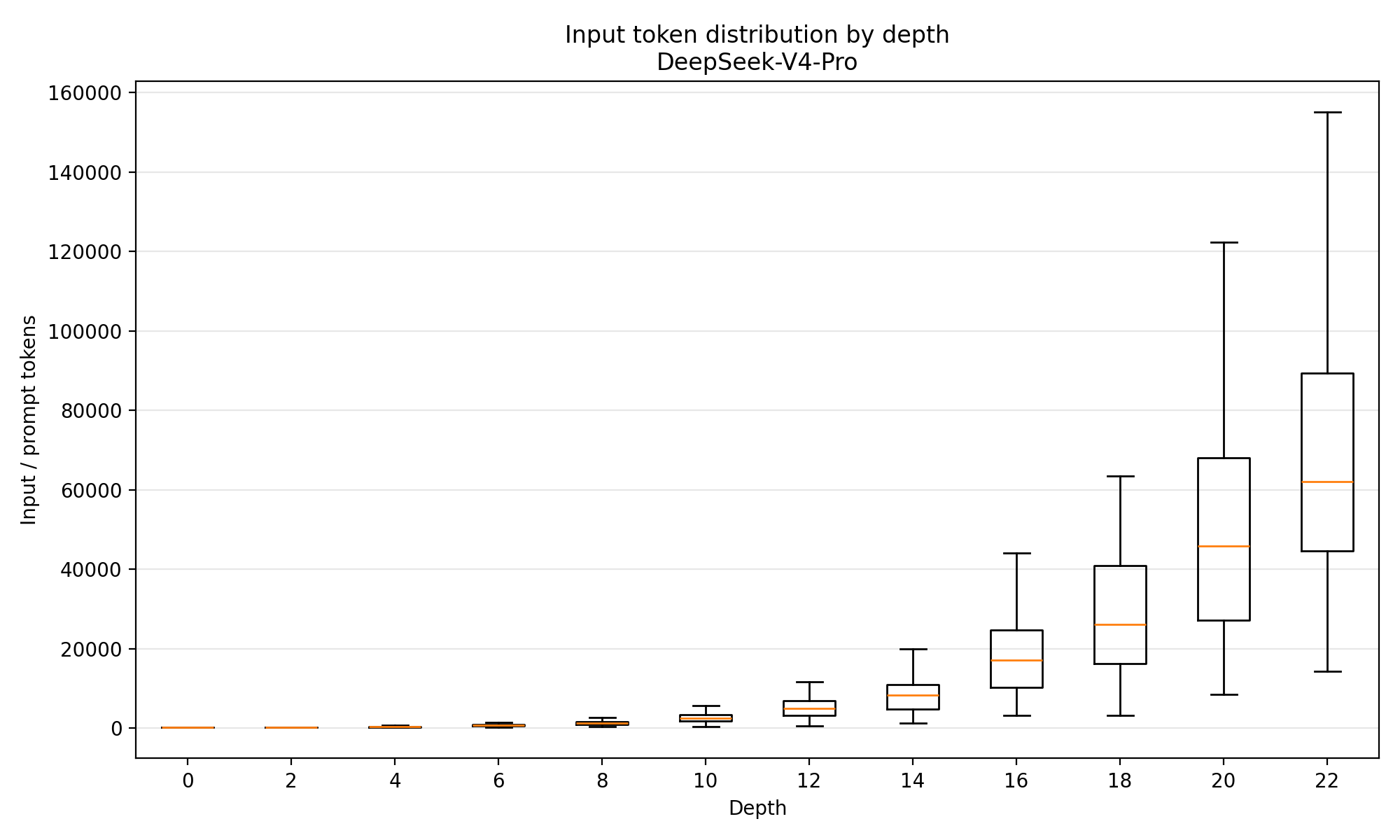}
  \caption{Boxplots for input tokens per Depth for \texttt{DeepSeek-V4-Pro}}
  \label{fig:input-tokens-boxplots}
\end{figure}
\begin{figure}[t]
\centering
\small
\begin{tikzpicture}[
    every node/.style={
        circle,
        draw,
        minimum size=7mm,
        inner sep=0pt
    },
    edge/.style={-{Latex[length=2.5mm]}, thick}
]

\node (p) at (0,0) {$p/2$};
\node (q) at (-1,-1) {$q/2$};
\node (s) at (1,-1) {$s/1$};
\node (r) at (-2,-2) {$r/1$};
\node (t) at (-0.15,-2) {$t/1$};
\node (u) at (-0.15,-3.25) {$u/1$};
\node (v) at (-0.15,-4.5) {$v/1$};

\draw[edge] (q) -- (p);
\draw[edge] (s) -- (p);

\draw[edge] (r) -- (q);
\draw[edge] (t) -- (q);

\draw[edge] (u) -- (t);
\draw[edge] (v) -- (u);
\end{tikzpicture}
\caption{Dependency graph of the program example in Figure \ref{fig:example}.}
\label{fig:context-dag}
\end{figure}
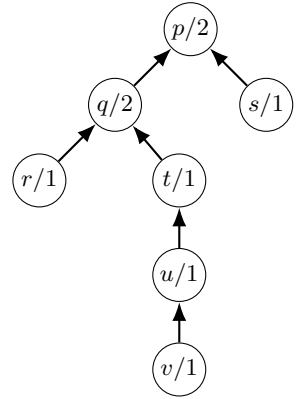
The benchmark statistics are shown in Table~\ref{tab:median-stats-by-depth}. The size of the context grows significantly with depth, Figure~\ref{fig:input-tokens-boxplots} shows boxplots of the input tokens as measured by \texttt{DeepSeek-V4-Pro}. Note that they are similar across the board, with a slight increase in \texttt{Claude-Opus-4.7}. Figure~\ref{fig:context-dag} shows the context DAG for the example in Figure~\ref{fig:example}.

\begin{table}[ht]
\centering
\footnotesize
\setlength{\tabcolsep}{3.5pt} 
\begin{tabular}{@{}rrrrrr@{}}
\hline
\textbf{depth} &
\textbf{rules} &
\textbf{facts} &
\textbf{preds} &
\begin{tabular}{c}\textbf{unique}\\\textbf{shapes}\end{tabular} &
\begin{tabular}{c}\textbf{tokens}\\\textit{(deepseek)}\end{tabular} \\
\hline
0  & 0.0    & 2.0    & 1.0    & 2   & 133 \\
2  & 3.0    & 9.0    & 7.0    & 77  & 197 \\
4  & 9.0    & 26.0   & 21.0   & 100 & 351 \\
6  & 23.0   & 66.0   & 56.5   & 100 & 690 \\
8  & 51.0   & 138.0  & 118.5  & 100 & 1329 \\
10 & 95.0   & 268.0  & 228.5  & 100 & 2543 \\
12 & 202.5  & 544.0  & 474.5  & 100 & 4939 \\
14 & 319.5  & 869.0  & 754.5  & 100 & 8319 \\
16 & 666.0  & 1788.0 & 1556.5 & 100 & 17059 \\
18 & 1052.5 & 2814.0 & 2457.5 & 100 & 26037 \\
20 & 1820.0 & 4855.0 & 4255.0 & 100 & 45870 \\
22 & 2424.5 & 6381.0 & 5604.0 & 100 & 62071 \\
\hline
\end{tabular}
\caption{Median \textsc{ProloNg} data statistics by depth.}
\label{tab:median-stats-by-depth}
\end{table}

We analyze the proofs derived by a symbolic solver, SWI-Prolog. Tables \ref{tab:provable_depth_time_trace} and \ref{tab:unprovable_depth_time_trace} show the median time and reasoning trace of SWI-Prolog for the provable and unprovable queries by depth. 
\begin{table}[ht]
\centering
\footnotesize

\begin{tabular}{rrrr}
\hline
\textbf{depth} & \textbf{n\_examples} & \textbf{time (s)} & \textbf{trace\_len} \\
\hline
0  & 100 & 0.03 & 3.0 \\
2  & 100 & 0.03 & 21.0 \\
4  & 100 & 0.03 & 63.0 \\
6  & 100 & 0.03 & 169.5 \\
8  & 100 & 0.04 & 355.5 \\
10 & 100 & 0.05 & 685.5 \\
12 & 100 & 0.07 & 1423.5 \\
14 & 100 & 0.08 & 2263.5 \\
16 & 100 & 0.13 & 4669.5 \\
18 & 100 & 0.19 & 7372.5 \\
20 & 100 & 0.34 & 12765.0 \\
22 & 100 & 0.42 & 16812.0 \\
\hline
\end{tabular}
\caption{Median time and trace length by depth for provable queries.}
\label{tab:provable_depth_time_trace}
\end{table}

\begin{table}[ht]
\centering
\footnotesize
\begin{tabular}{rrrr}
\hline
\textbf{depth} & \textbf{n\_examples} & \textbf{time (s)} & \textbf{trace\_len} \\
\hline
0  & 100 & 0.03 & 2.0 \\
2  & 100 & 0.03 & 14.0 \\
4  & 100 & 0.03 & 32.0 \\
6  & 100 & 0.03 & 80.0 \\
8  & 100 & 0.04 & 177.5 \\
10 & 100 & 0.05 & 353.0 \\
12 & 100 & 0.07 & 635.0 \\
14 & 100 & 0.08 & 920.0 \\
16 & 100 & 0.12 & 1952.0 \\
18 & 100 & 0.18 & 3278.0 \\
20 & 100 & 0.31 & 4412.0 \\
22 & 100 & 0.41 & 6183.5 \\
\hline
\end{tabular}
\caption{Median time and trace length by depth for unprovable queries.}
\label{tab:unprovable_depth_time_trace}
\end{table}

\section{Accuracy by provability}
\begin{figure*}[ht]
  \includegraphics[width=\textwidth]{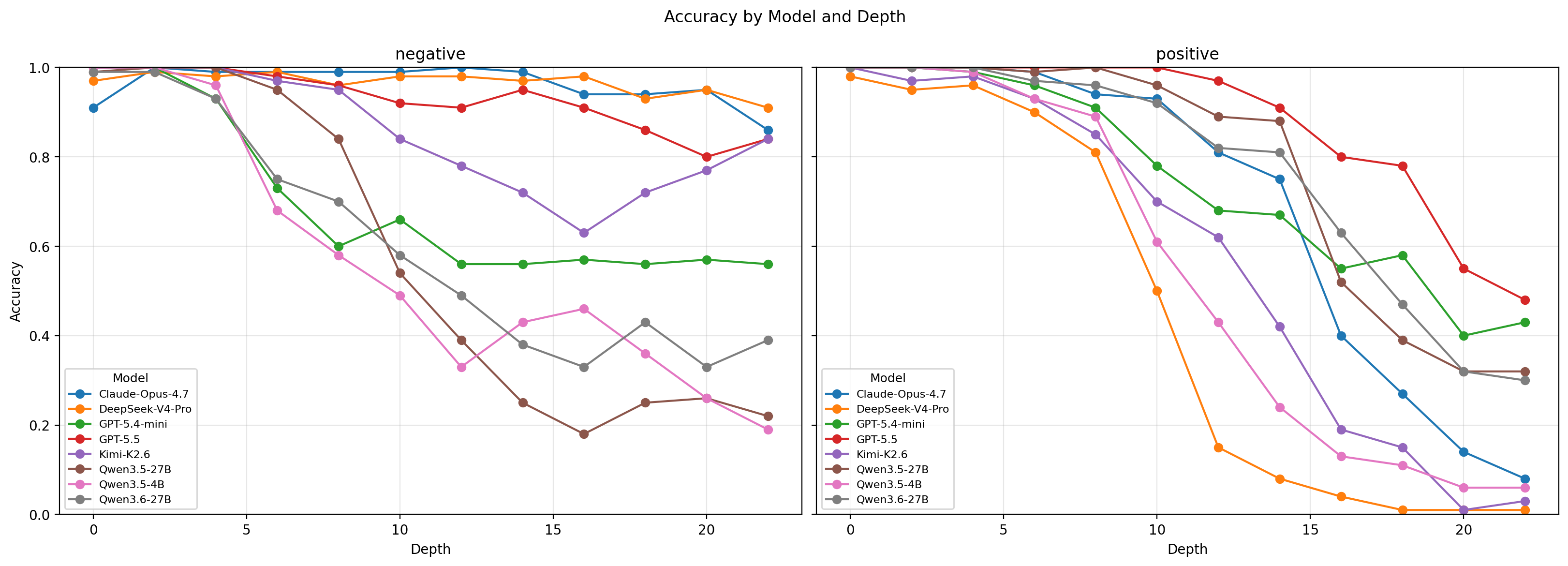}
  \caption{Accuracy per target provability. The deterioration in performance is sharper for provable queries as complexity increases, which aligns with the models' \textit{conservative} bias. }
  \label{fig:acc-per-target}
\end{figure*}

We show in Figure~\ref{fig:acc-per-target} how the accuracy is affected by depth divided by whether or not a query is provable or not. We can see that the deterioration is less drastic in unprovable queries.

\section{Growth of reasoning trace with complexity}
\label{sec:trace-len}

\begin{figure}[h]
  \includegraphics[width=\columnwidth]{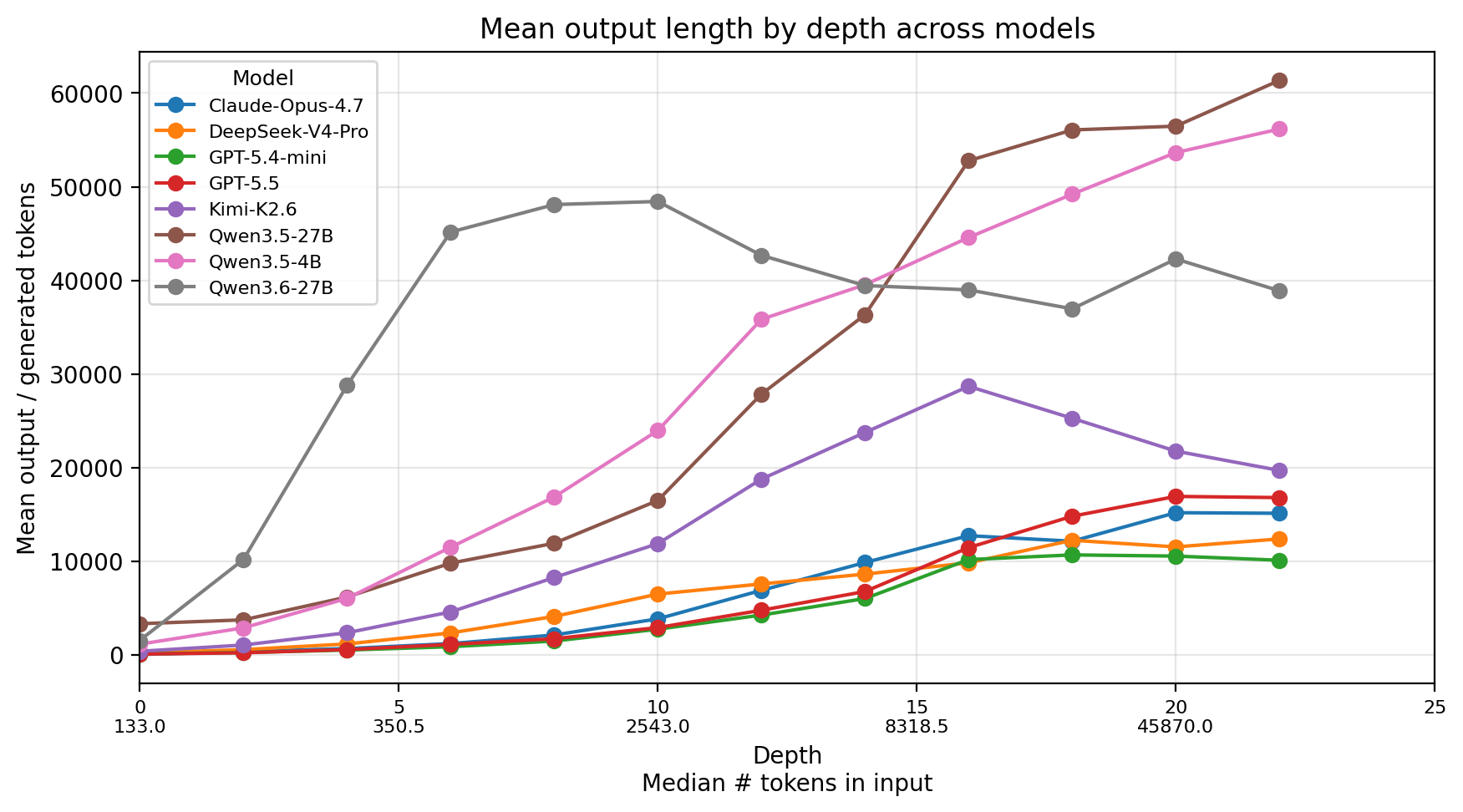}
  \caption{Mean number of output tokens per model. Only \texttt{Qwen-3.5} continues to grow monotonically, with the rest remaining significantly below the token budget.}
  \label{fig:mean-tokens-per-depth}
\end{figure}

The growth of the reasoning trace varies across models, as shown in Figure~\ref{fig:mean-tokens-per-depth}. 
\texttt{GPT-5.5}, \texttt{Claude-Opus-4.7}, \texttt{DeepSeek-V4-Pro}, and \texttt{GPT-5.4-mini} follow broadly similar trends, generating fewer than $20k$ output tokens across all depths with only modest increases as depth grows, suggesting these models allocate reasoning budget efficiently regardless of problem complexity. \texttt{Qwen3.6-27B} displays a notably irregular pattern: output length peaks around depths $8-10$ before declining to a plateau of approximately $40k$ tokens at greater depths (note that the maximum budget was $82k$). \texttt{Kimi-K2.6} similarly exhibits a non-monotonic profile, with output length peaking around depth 16 before declining. \texttt{Qwen3.5} stands apart from all other models in its tendency to allocate monotonically increasing reasoning tokens with depth, with output length continuing to grow even at the deepest levels, a pattern consistent with the model expanding its thinking trace without necessarily improving its derivations.
\section{Mixed-Effects Model Details}
\label{app:glmm}

For all regression analyses in Section~\ref{sec:disentangle}, we fit
generalized linear mixed models (GLMMs) with a binomial family and
logit link, using \texttt{lme4::glmer} in R \cite{bates2015lme4}, with
the \texttt{bobyqa} optimizer. Model identity is included as a random
intercept, $(1 \mid \text{model\_id})$, rather than a fixed effect, to
account for baseline differences in model capability without assuming
a specific reference model. Context length (log-transformed) and
proof depth are included as fixed effects, standardized (z-scored)
prior to fitting.

\paragraph{Original data, all models ($k=8$).} Fit on the full,
non-disentangled evaluation set across all eight models
($n=19{,}200$), the random-intercept variance for model identity was
$\hat\sigma^2 = 0.607$ ($\text{SD}=0.779$). Context length showed a
large, significant effect ($\beta=-1.38$, $p<0.001$), while proof
depth's effect was much smaller
($\beta=-0.18$, $p=0.012$), consistent with the confounding described
in Section~\ref{sec:disentangle}.

\paragraph{Original data, matched models ($k=4$).} Restricting to the
same four models used in the disentangled experiment
($n=9{,}600$), to rule out model composition as a confound, yields a
smaller random-intercept variance ($\hat\sigma^2=0.156$,
$\text{SD}=0.395$) and a similar pattern: context length remains large
and significant ($\beta=-1.48$, $p<0.001$), while proof depth is no
longer statistically significant ($\beta=-0.15$, $p=0.108$).

\paragraph{Disentangled data ($k=4$).} On the disentangled dataset
($n=2{,}400$), the random-intercept variance is $\hat\sigma^2=0.347$
($\text{SD}=0.589$), and both fixed effects are significant and
comparable in magnitude (context length: $\beta=-0.54$, $p<0.001$;
proof depth: $\beta=-0.36$, $p<0.001$).

\paragraph{Per-model random intercepts.} Table~\ref{table:blups}
reports the estimated random intercept
for each of the four models in the original and disentangled settings, representing each model's
deviation in baseline log-odds of correctness from the average across
models. 
\begin{table}[t]
\centering
\small
\begin{tabular}{lcc}
\toprule
\textbf{Model} & \textbf{Original} & \textbf{Disentangled} \\
\midrule
\texttt{Qwen3.5-4B}      & $-0.668$ & $-0.861$ \\
\texttt{Qwen3.5-27B}     & $ 0.127$ & $ 0.744$ \\
\texttt{Qwen3.6-27B}     & $ 0.226$ & $ 0.189$ \\
\texttt{DeepSeek-V4-Pro} & $ 0.315$ & $-0.083$ \\
\bottomrule
\end{tabular}
\caption{Per-model random intercept estimates (BLUPs) from the GLMM,
comparing the matched (k=4) original and disentangled settings.} 
\label{table:blups}
\end{table}
\section{Prompt Template}
\label{sec:appendix}
\begin{tcolorbox}[
  colback=purple!5!white,
  colframe=purple!55!black,
  title=\textbf{System Prompt},
  fonttitle=\bfseries, 
  breakable
]
\begin{lstlisting}
You are a helpful and harmless assistant. Consider the following prolog-like declarative logic program, arguments starting with capital letters represent variables, lowercase letters represent constants. \+ is a symbol that represents negation by failure, a negated literal in a rule is satisfied when it fails to be proven. Decide if a query can be proven given the background program and return 1, otherwise return 0 if it cannot be proven. A query is the final line starting with ""?- "". You should think step-by-step. Put your final answer within \\boxed{{}}.
\end{lstlisting}
\end{tcolorbox}
\section{Full Example}
\label{appendix:ex}

\begin{tcblisting}{
  colback=gray!5!white,
  colframe=gray!55!black,
  title=\textbf{Qwen-3-Thinking Input},
  fonttitle=\bfseries,
  listing only,
  breakable,
  listing options={
    basicstyle=\small\ttfamily,
    breaklines=true,
    columns=fullflexible,
    aboveskip=0pt, 
    belowskip=0pt
  }
}
<|im_start|>system
You are a helpful and harmless assistant. Consider the following prolog-like declarative logic program, arguments starting with capital letters represent variables, lowercase letters represent constants. \+ is a symbol that represents negation by failure, a negated literal in a rule is satisfied when it fails to be proven. Decide if a query can be proven given the background program and return 1, otherwise return 0 if it cannot be proven. A query is the final line starting with ""?- "". You should think step-by-step. Put your final answer within \\boxed{{}}.<|im_end|>
<|im_start|>user
p(X,Y):- q(Y,X), s(X).
q(X,Y):- r(X), t(Y).
t(X):- u(X).
u(X):- v(X).
r(o1).
r(o2).
s(o1).
v(o1).
?- p(o1, o2).<|im_end|>
<|im_start|>assistant
<think>
\end{tcblisting}

\begin{tcolorbox}[
  colback=yellow!5!white,
  colframe=yellow!55!black,
  title=\textbf{Qwen-3-30B Thinking Reasoning Content},
  fonttitle=\bfseries,
  breakable
]
\begin{lstlisting}
First, I need to figure out if the query ""?- p(o1, o2)."" can be proven based on the given program. The program has several rules and facts. Let me list them out clearly.

The program is:

- p(X,Y) :- q(Y,X), s(X).

- q(X,Y) :- r(X), t(Y).

- t(X) :- u(X).

- u(X) :- v(X).

- r(o1).

- r(o2).

- s(o1).

- v(o1).

The query is p(o1, o2).

I need to see if p(o1, o2) can be proven using these rules. Since it's a declarative logic program, I should try to derive it step by step.

Start with the query: p(o1, o2).

From the first rule, p(X,Y) is true if q(Y,X) and s(X) are both true.

So for p(o1, o2), X is o1 and Y is o2, so it requires q(o2, o1) and s(o1).

Now, s(o1) is given as a fact: s(o1). So s(o1) is true. That part is satisfied.

Now, I need to check if q(o2, o1) can be proven.

q(X,Y) is true if r(X) and t(Y) are both true.

For q(o2, o1), X is o2, Y is o1, so it requires r(o2) and t(o1).

r(o2) is given as a fact: r(o2). So r(o2) is true.

Now, I need t(o1).

t(X) is true if u(X) is true.

u(X) is true if v(X) is true.

v(o1) is given as a fact: v(o1). So v(o1) is true.

Therefore, u(o1) should be true because u(X) :- v(X), so with X=o1, u(o1) is true.

Then t(o1) should be true because t(X) :- u(X), so with X=o1, t(o1) is true.

Therefore, t(o1) is true.

So for q(o2, o1), we have r(o2) true and t(o1) true, so q(o2, o1) is true.

Then, for p(o1, o2), we have q(o2, o1) true and s(o1) true, so p(o1, o2) should be true.

Let me double-check.

List all the steps:

1. Query: p(o1, o2)

2. From p(X,Y) :- q(Y,X), s(X), so for X=o1, Y=o2: need q(o2, o1) and s(o1)

3. s(o1) is fact, so s(o1) = true

4. Now need q(o2, o1)

5. From q(X,Y) :- r(X), t(Y), so for X=o2, Y=o1: need r(o2) and t(o1)

6. r(o2) is fact, so r(o2) = true

7. Now need t(o1)

8. From t(X) :- u(X), so for X=o1: need u(o1)

9. From u(X) :- v(X), so for X=o1: need v(o1)

10. v(o1) is fact, so v(o1) = true

11. Therefore u(o1) = true (from u(X):-v(X))

12. Therefore t(o1) = true (from t(X):-u(X))

13. Therefore q(o2, o1) = true (from q(X,Y):-r(X),t(Y) with X=o2,Y=o1)

14. Therefore p(o1, o2) = true (from p(X,Y):-q(Y,X),s(X) with X=o1,Y=o2)

All steps seem to check out. But let me make sure there are no negations or anything that might cause failure.

The program has \+ for negation by failure, but in this case, there are no negated literals in the rules or the query. The rules are all positive implications.

For example, the rule for p has q(Y,X) and s(X), both positive.

q has r(X) and t(Y), positive.

t has u(X), positive.

u has v(X), positive.

Facts are given directly.

So no negation to worry about.

Also, the variables: in the query, o1 and o2 are constants, not variables, so we're directly substituting.

In the rule, X and Y are variables, but in the query, we're using specific constants.

So it should be fine.

I should confirm if all the facts are present.

r(o1) and r(o2) are both facts, so r is true for o1 and o2.

s(o1) is fact, so s is true for o1.

v(o1) is fact, so v is true for o1.

Now, for t(o1), we need u(o1), which needs v(o1), which is true.

Similarly, for q(o2,o1), we need r(o2) which is true, and t(o1) which we just established.

r(o2) is true, yes.

Is there any other path? I think this is the only way.

The query is p(o1,o2), and we've derived it.

But let me see the rule for p: p(X,Y) :- q(Y,X), s(X).

So for p(o1,o2), it requires q(Y,X) with Y=o2, X=o1, so q(o2,o1).

Yes.

In q(X,Y), it's r(X) and t(Y), so for q(o2,o1), X=o2, Y=o1, so r(o2) and t(o1).

All good.

I could think in terms of the body.

The body of p(o1,o2) is q(o2,o1) and s(o1).

s(o1) is true.

q(o2,o1) requires r(o2) and t(o1).

r(o2) is true.

t(o1) requires u(o1), which requires v(o1), which is true.

So yes.

The program also has r(o1), but we didn't use it for this query, which is fine.

Similarly, s(o1) is used, but s(o2) isn't given, so we don't care.

v(o1) is used, but v(o2) isn't given, so not needed.

So for this specific query, it should be provable.

Therefore, the answer should be 1.

The question says ""return 1 if it can be proven, otherwise 0"".

So I think it can be proven.

But let me make sure about the negation by failure. The problem mentions that \+ represents negation by failure, but in this program, there is no \+ used. So no issue.

For example, if there was a rule like p(X) :- \+ q(X), but here there isn't.

So all good.

Perhaps I should write it out as a proof tree.

Query: p(o1, o2)

- p(o1, o2) requires q(o2, o1) and s(o1)

- s(o1) is true (fact)

- q(o2, o1) requires r(o2) and t(o1)

- r(o2) is true (fact)

- t(o1) requires u(o1)

- u(o1) requires v(o1)

- v(o1) is true (fact)

Therefore, all subgoals are satisfied, so p(o1, o2) is true.

Hence, it can be proven.

So the answer should be 1.

The problem says ""arguments starting with capital letters represent variables, lowercase letters represent constants."" Here, o1 and o2 are lowercase, so constants, which is fine.

In the rule, X and Y are variables, but we're instantiating them.

No issue.

I think I'm confident.
</think>

\end{lstlisting}
\end{tcolorbox}
\begin{tcolorbox}[
  colback=green!5!white,
  colframe=green!55!black,
  title=\textbf{Qwen-3-30B Thinking Answer},
  fonttitle=\bfseries,
  breakable
]
\begin{lstlisting}
To determine if the query `?- p(o1, o2).` can be proven based on the given program, we need to check if the rules and facts allow us to derive `p(o1, o2)` through logical inference. The program is declarative, and negation by failure (`\+`) is not present in any rules or the query, so we only deal with positive literals.

### Step-by-Step Derivation:
1. **Query: `p(o1, o2)`**
   - From the rule `p(X, Y) :- q(Y, X), s(X)`, we substitute `X = o1` and `Y = o2`:
     - This requires `q(o2, o1)` and `s(o1)` to be true.

2. **Check `s(o1)`**:
   - `s(o1)` is given as a fact in the program: `s(o1).`
   - Therefore, `s(o1)` is true.

3. **Check `q(o2, o1)`**:
   - From the rule `q(X, Y) :- r(X), t(Y)`, we substitute `X = o2` and `Y = o1`:
     - This requires `r(o2)` and `t(o1)` to be true.

4. **Check `r(o2)`**:
   - `r(o2)` is given as a fact in the program: `r(o2).`
   - Therefore, `r(o2)` is true.

5. **Check `t(o1)`**:
   - From the rule `t(X) :- u(X)`, we substitute `X = o1`:
     - This requires `u(o1)` to be true.
   - From the rule `u(X) :- v(X)`, we substitute `X = o1`:
     - This requires `v(o1)` to be true.
   - `v(o1)` is given as a fact in the program: `v(o1).`
     - Therefore, `v(o1)` is true.
   - Since `v(o1)` is true, `u(o1)` is true (from `u(X) :- v(X)`).
   - Since `u(o1)` is true, `t(o1)` is true (from `t(X) :- u(X)`).

6. **Conclusion for `q(o2, o1)`**:
   - `r(o2)` is true and `t(o1)` is true, so `q(o2, o1)` is true.

7. **Conclusion for `p(o1, o2)`**:
   - `q(o2, o1)` is true and `s(o1)` is true, so `p(o1, o2)` is true.

All subgoals required to prove `p(o1, o2)` are satisfied by the facts and rules. There are no negated literals or failures that prevent the proof.

### Final Answer:
Since the query `p(o1, o2)` can be proven, the answer is 1.
\boxed{1}
\end{lstlisting}
\end{tcolorbox}
\end{document}